\documentclass[unnumsec,webpdf,contemporary,large]{oup-authoring-template}%
\usepackage{verbatim}
\usepackage{threeparttable}
\usepackage{bm}
\usepackage{romanbar}

\graphicspath{{Fig/}}

\theoremstyle{thmstyleone}%
\theoremstyle{thmstyletwo}%
\theoremstyle{thmstylethree}%

\begin{document}

\journaltitle{Briefings in Bioinformatics}
\DOI{DOI HERE}
\copyrightyear{2022}
\pubyear{2022}
\access{}
\appnotes{Problem Solving Protocols}

\firstpage{1}


\title[Short Article Title]{TCR: A Transformer Based Deep Network for Predicting Cancer Drugs Response}

\author[1]{Jie Gao}
\author[1]{Jing Hu}
\author[1]{Wanqing Sun}
\author[1]{Yili Shen}
\author[2]{Xiaonan Zhang}
\author[2, $\ast$]{Xiaomin Fang}
\author[2, $\ast$]{Fan Wang}
\author[1, $\ast$]{Guodong Zhao}
\authormark{Author Name et al.}

\address[1]{\orgname{Baidu, Inc.}, \orgaddress{\street{701, Na Xian Road}, \postcode{201210}, \state{Shanghai}, \country{China}}}
\address[2]{\orgname{Baidu, Inc.}, \orgaddress{\street{Xue Fu Road}, \postcode{518000}, \state{Shenzhen}, \country{China}}}


\corresp[$\ast$]{Corresponding author. \href{email:email-id.com}Email:{ gaojie01@baidu.com}, {zhaoguodong@baidu.com}, {fangxiaomin01@baidu.com}, {wangfan04@baidu.com}}




\abstract{
Predicting clinical outcomes to anti-cancer drugs on a personalized basis is challenging in cancer treatment due to the heterogeneity of tumors. Traditional computational efforts have been made to model the effect of drug response on individual samples depicted by their molecular profile, yet overfitting occurs because of the high dimension for omics data, hindering models from clinical application. Recent research shows that deep learning is a promising approach to build drug response models by learning alignment patterns between drugs and samples. However, existing studies employed the simple feature fusion strategy and only considered the drug features as a whole representation while ignoring the substructure information that may play a vital role when aligning drugs and genes. Hereby in this paper, we propose TCR (Transformer based network for Cancer drug Response) to predict anti-cancer drug response. By utilizing an attention mechanism, TCR is able to learn the interactions between drug atom/sub-structure and molecular signatures efficiently in our study. Furthermore, a dual loss function and cross sampling strategy were designed to improve the prediction power of TCR. We show that TCR outperformed all other methods under various data splitting strategies on all evaluation matrices (some with significant improvement). Extensive experiments demonstrate that TCR shows significantly improved generalization ability on independent in-vitro experiments and in-vivo real patient data. Our study highlights the prediction power of TCR and its potential value for cancer drug repurpose and precision oncology treatment.}

\maketitle
\section{Introduction}
Predicting clinical response to anti-cancer drugs is crucial in cancer treatment. Cancer is a multifactorial and heterogeneous disease. Hence, personalized, instead of one-size-fits-all, approaches to therapy are considered to have great potential to improve the clinical treatment of cancer~\citep{turki2017link}.

Fortunately, large-scale projects on drug response for “artificial patients” (i.e., cell line), such as GDSC~\citep{yang2012genomics}, CCLE~\citep{ghandi2019next}, and PRISM~\citep{yu2016high}, have promoted the development of computational methods for drug response prediction ~\citep{azuaje2017computational,chen2021survey,firoozbakht2022overview}.

Besides, the release of high-throughput sequencing data of cancer cell lines enabled the study of the cancer phenotype using multi-omics data, such as genomic, transcriptomic, and epigenomics~\citep{gagan2015next}. 
For instance, CCLE (Cancer Cell Line Encyclopedia) has published around 1000 cancer cell lines and their corresponding omics profiles~\citep{ghandi2019next, nusinow2020quantitative}. The half-maximal inhibitory concentration ($IC_{50}$) scores have been analyzed to identify the drug sensitivity across cancer cell lines.
Based on the above data, many research teams have launched an impact on the challenge of drug sensitivity prediction. Most of them were machine learning-based, where different data and model strategies were introduced. One main category of these methods was traditional machine learning-based methods. Elastic-net was the model previously applied on GDSC to identify features associated with drug response~\citep{garnett2012systematic}.
In~\citep{wan2014ensemble}, the random forest regression model was used for obtaining a single drug or a group of drugs for specific cancer. Besides, ensemble learning strategies were used to integrate individual models for accurate prediction~\citep{matlock2018investigation, tan2019drug}. 
However, today’s complex omics datasets have appeared too multidimensional to be effectively managed by traditional machine learning algorithms. 

Deep learning is a state-of-the-art branch of machine learning for extracting features from complex data and making accurate predictions. Recently, deep learning has been applied to drug response prediction and achieved superior performance compared to traditional machine learning methods~\citep{liu2019improving,  nguyen2021graph, liu2020deepcdr}.
Deep learning can be utilized to automatically learn latent features of cell lines and the structural features of drugs to predict anti-cancer drug responsiveness. For example, tCNNs~\citep{liu2019improving} applied a CNN for predicting cancer drug response (CDR) using simplified molecular input line entry specification (SMILES) sequences of drugs and genomic mutation data of cancer cell lines. SMILES sequences of drugs are encoded into one-hot representation and fed to the neural network. GraphDRP~\citep{nguyen2021graph} used a graph convolutional network (GCN) to capture the structural features of molecules. A fully connected layer (FC layer) followed by GCN converted the result to 128 dimensions representing the whole drug. Meanwhile, features of cell lines were extracted by convolution layers. Then combined features of drugs and cell lines were used to represent each drug-cell line pair. DeepCDR~\citep{liu2020deepcdr} proposed a hybrid graph convolutional network that integrated multi-omics profiles and explored intrinsic chemical structures of drugs for predicting CDR. The high-level features of drugs and multi-omics data were then simply concatenated together and fed into a 1D-CNN to predict the $IC_{50}$ sensitivity values.

However, these above mentioned methods have some limitations as follows.
\begin{itemize}
\item The comparative methods only considered the drug features as a whole representation while ignoring the substructure information that may play a vital role when aligning drugs and genes~\citep{fotis2021deepsiba}. Thus, simply concatenating the whole drug and omics features may lose focus on the meaningful substructure information that are related to compounds’ biological activity~\citep{klekota2008chemical,pope2018discovering}.

\item Most existing methods use overall PCC (PCC across all drugs) as a metric in multi-drug prediction scenarios, which can not measure the model's performance correctly.


\end{itemize}


We present TCR for cancer drugs response prediction:

1.We proposed a TCR model, in which an attention mechanism was designed to fuse the drug and omics features and capture the different interactions among drug sub-structure and multi-omics data that may indicate the MoA of drugs. Furthermore, we designed a dual loss function and cross sampling strategy to improve the prediction.

2. We proposed a new evaluation metric per-drug Pearson correlation coefficient named DrugPCC, which can better measure the model's performance than the previous overall PCC index.

3. We conducted extensive experiments (from cell line data to clinical data) to identify the generalization of our model. TCR achieved state-of-the-art performance.


\section{Materials and methods}\label{sec2}

\begin{figure*}[h!]
\centering
\includegraphics[width=1\textwidth]{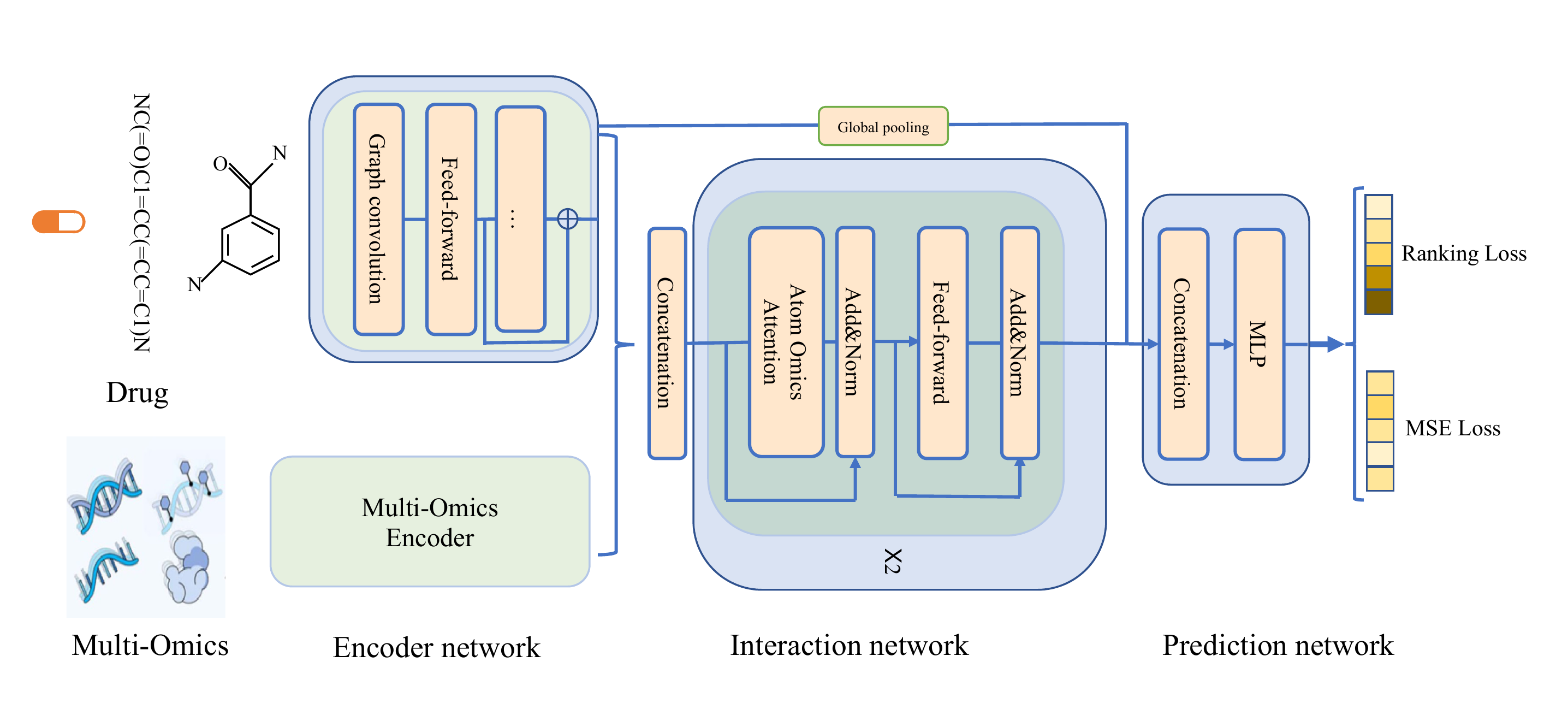}
\caption{ The proposed TCR model. Three components construct our TCR model: an encoder module that encodes drugs and multi-omics features, an interaction module that models the interactions of drug atoms and omics, and a prediction module that outputs $IC_{50}$ values (natural log-transformed).}
\label{fig:01}
\end{figure*}

The proposed model is presented in \textbf{Fig.~\ref{fig:01}}. 
Three components construct our TCR model: An encoder module that encodes drugs and multi-omics inputs. For drugs, we use graph convolution network~\citep{kipf2016semi} as the backbone to learn the structure feature of drugs by taking the adjacent information of atoms into consideration. The atom features of the drug are aggregated from neighboring atoms. The input of each drug features are the attributes of each atom in a compound including chemical and topological properties such as atom type, degree and hybridization~\citep{ramsundar2019deep}.
For omics, three omics-specific subnetworks is designed to encode the information of genomic, transcriptomic, and epigenomic.

An interaction module that models the interactions of drug substructures and omics information.
The main structure of the interaction module is a transformer network. It contains a multi-head atom omics attention layer, a feed-forward propagation layer. Add-norm layers are used between the previous two layers. For a detailed interaction module structure, see~\ref{Interaction}

$IC_{50}$ values, which denotes the effectiveness of a drug in inhibiting the growth of a specific cancer cell line, is predicted by a prediction module. A small $IC_{50}$ value reveals a high degree of drug efficacy, implying that the cancer cell line is sensitive to the corresponding drug. 

Furthermore, ranking loss~\citep{chen2009ranking, cao2007learning} and regression loss are used in our method. For the prediction of cancer drug response, we not only regard it as a regression problem but also as the ranking of effectiveness.
We first used MSE as the regression loss function. Then, in order to fit the actual scenario: we payed more attention to the drug effectiveness ranking for a sample rather than the actual $IC_{50}$ absolute values. 
We designed a pair-wise ranking~\citep{wang2019ranked} loss to focus on the effectiveness of the same drug in different individuals and the effectiveness ranking of all drugs in the same sample. 
We converted the ground truth $IC_{50}$ values into a ranking list, and the model was applied to predict the pair-wise rank of samples. We designed a cross-sampling method to cooperate with ranking loss and mine hard examples. For a detailed description of losses, see~\ref{loss}

\subsection{Data preparation}

In our study, we used four public datasets: GDSC ~\citep{iorio2016landscape}, 
CCLE~\citep{barretina2012cancer}, PRISM~\citep{yu2016high}, and TCGA(The Cancer Genome Atlas)~\citep{weinstein2013cancer}. GDSC is a large-scale drug screening dataset which provides $IC_{50}$ values for each drug-cancer cell line pair. CCLE provides genomic, transcriptomic, and epigenomic profiles for more than 1,000 cancer cell lines. We used genomic mutation, transcriptomic (gene expression), and DNA methylation data as multi-omics inputs, which can be easily accessed and downloaded using the DepMap portal (https://depmap.org). 
PRISM and TCGA patient data are used for external validation. 
The primary PRISM Repurposing dataset contains the results of pooled-cell line chemical-perturbation viability screens for 4,518 compounds screened against 578 or 562 cell lines. PRISM is a molecular barcoding method to screen drugs against cell lines in pools. In this work, we only used PRISM drug screening data to evaluate our model's generalization on cell line data.
Furthermore, we used TCGA data to verify the performance of our model in clinical scenes. TCGA patient dataset provides genetic profiles of patients and clinic annotation for over 11,000 patients across 33 cancer types. The TCGA dataset can be downloaded from Genomic Data Commons Data Portal.

We followed the same data preparation processing of DeepCDR~\citep{liu2020deepcdr}. The $IC_{50}$ values downloaded from the GDSC database were natural log-transformed and utilized as the ground truth. Drug-cell line pairs without PubChem ID or omics data were removed. We finally obtained a dataset containing 94,076 samples consisting of 493 cancer cell lines and 223 drugs. 

For multi-omics profiles of cancer cell lines, we only considered omics associated with 697 genes from COSMIC Cancer Gene Census (https://cancer.sanger.ac.uk/census)~\citep{forbes2006cosmic}. For transcriptomic data, the TPM values of gene expression were log2 transformed and quantile normalized. Then the transcriptomic (gene expression) value of each cell line could be represented as a 697-dimensional feature vector. For genomic mutation, 34,673 unique genomic positions, including SNPs and Indels in the above genes, were collected. The genomic mutation of each cancer cell line was represented as a binary feature vector in which '1' denotes a mutated position, and '0' denotes a non-mutated position. The DNA methylation data were obtained from the processed Bisulfite sequencing data~\citep{krueger2012dna} of promoters 1kb upstream TSS region. A median value interpolation method was applied to the data as there were a minority of missing values. 

We used the clinical data of drugs with large sample numbers from The Cancer Genome Atlas (TCGA) for verification. Following ~\citep{ding2016evaluating}, we collected six drugs: Temozolomide, Gemcitabine, Cisplatin, Fluorouracil, Carboplatin, and Paclitaxel. The number of samples for each drug is larger than 100. We treated partial and complete responses as responders and progressive clinical disease and stable disease as non-responders.

\begin{methods}
\section{Methods}
\subsection{Encoder modules}
We defined an encoder module for drug and multi-omics features extraction. 

For drug encoder module, each drug has its unique chemical topological structure, which can be naturally represented as a graph where the nodes and edges represent chemical atoms and bonds. For $M$ drugs, we have $M$ graphs described as ${{G_{i} = (X_{i},A_{i})|_{i=1}^M}}$ where $\mathbf{X}_{i} \in \mathbb{R}^{N_{i} \times C}$ is the feature matrix of atoms of $i$th drug and $C$ is the feature dimension of each atom, $N_{i}$ is the number of atoms in the $i$th drug. By following the illustration in~\citep{ramsundar2019deep}, we set $C = 75$. Then, each atom's attributes, including chemical and topological properties, were represented as a 75-dimension vector. Moreover, this vector was used as the initial feature of each atom. We downloaded the structure files of all drugs from PubChem library~\citep{kim2019pubchem}.

As shown in \textbf{Fig.~\ref{fig:01}}, different from previous works that aim to achieve graph-level features of drugs, we aim to obtain the node/atom level features of the drug, which was then used to model atom-level interactions with omics data. The backbone of the drug encoder network is GCN~\citep{kipf2016semi}. We padded zeros in drug feature $X$ for those drugs whose number of atoms is less than 100. After padding, the atom features and adjacent matrix are represented as $\mathbf{X}_{i}^{\prime} \in \mathbb{R}^{N \times C}$, $\mathbf{A}_{i}^{\prime} \in \mathbb{R}^{N \times N}$ of the $i$th drug ,where $N = 100$.
The hidden state of each drug $H_{i}$ is layer-wise updated by the below operation:


\begin{equation}
\mathbf{H}_{i}^{(l+1)}=\sigma\left({{\mathbf{D}}}^{-\frac{1}{2}} {\mathbf{A}}_{i}^{\prime} {{\mathbf{D}}}^{-\frac{1}{2}} \mathbf{H}_{i}^{(l)} \boldsymbol{W}^{(l)}\right)
\end{equation}

where $\mathbf{A}_{i}^{\prime}$ is the adjacent matrix with self-connection. $\mathbf{H}_{i}^{(l)}$ is the $l$ layer's hidden state. ${W}^{(l)}$ is $l$ layer's trainable weight matrix. $\mathbf{D}$ is the degree matrix. Thus we got every atom representation of each drug. Besides, we also used a global pooling layer to get the graph level representation.

For multi-omics encoder network, three omics-specific subnetworks were designed to learn the desentangled representation of each omics data. Three seperated omic encoder networks representated as ${r_{g} = G_{g}(x_{g})}$, ${r_{t} = G_{t}(x_{t})}$, ${r_{e} = G_{e}(x_{e})}$ for processing genomic, transcriptomic, and epigenomic data. We directly used fully connected networks for feature representation of omics data.

\subsection{Interaction module}
\label{Interaction}
The interaction module models the interactions among omics data and drug substructure/atoms.
As shown in \textbf{Fig.~\ref{fig:01}}, a transformer block was applied to model the interactions among omics and drugs. 
Transformer is a type of neural network mainly based on the self-attention mechanism ~\citep{vaswani2017attention}, which can capture the relationships among different features. Transformer shows superior performance in the field of natural language processing (NLP) and computer vision (CV), e.g., the famous BERT~\citep{devlin2018bert}, GPT-3~\citep{brown2020language}, DETR~\citep{carion2020end} and Swin Transformer ~\citep{liu2021swin} models. As far as we know, TCR is the first model that applies the transformer network on CDR task. 
Instead of performing a single attention function,~\citep{vaswani2017attention} found it was beneficial to capture different contexts with multiple individual attention functions. We designed a multi-head attention module termed atom omics attention (AOA) to capture the associations among drug atom/nodes and omics. The encoded drug node level features $\mathbf{R_{drug}} \in \mathbb{R}^{N \times C}$ and omics features $\mathbf{R_{omics}} = [r_g,r_t,r_e]$ were concatenated and fed into interaction module.
The interaction module consists of 2 transformer blocks \textbf{Fig.~\ref{fig:01}}, each containing an atom omics attention (AOA) layer, a feed-forward layer, and two add-and-norm layers. 

\textbf{Fig.~\ref{fig:02}} shows the detailed structure of atom omics attention. Different from the self-attention mechanism, every drug atom query pays attention to different genomic, transcriptomic, and epigenomic features, respectively.

Attention function maps a sequence of query $ Q_{drug} = [{r_{drug}^{1}, r_{drug}^{2}. . . , r_{drug}^{N} }]$ and a set of key-value pairs  $ K_{omics}, V_{omics}  = [{(r_{g}, r_{g}),(r_{t}, r_{t}), (r_{e}, r_{e})}]$ to outputs. More specifically, multi-head attention model first transforms $Q_{drug}$, $K_{omics}$, and $V_{omics}$ into $h$ subspaces, with different learnable linear projections. Where $h$ is the number of heads. Furthermore, $h$ attention functions are applied parallel to produce the output states $H$, where $H_{i}$ represent the $i$th output state produced by
the $i$th attention head.
\begin{equation}
\operatorname{H}_{\mathrm{i}}=\operatorname{Attention}\left(Q_{drug}W_{i}^{Q},K_{omics}W_{i}^{K},V_{omics}W_{i}^{V}\right)
\end{equation}
\begin{equation}
\operatorname{Attention}(Q, K, V)=\operatorname{softmax}\left(\frac{Q K^{T}}{\sqrt{d_{k}}}\right) V
\end{equation}
Where the projections are parameter matrices 
$W_{i}^{Q}\in \mathbb{R}^{d_{\text{model}}\times d_{k}}$, 
$W_{i}^{K}\in \mathbb{R}^{d_{\text{model}}\times d_{k}}$ and 
$W_{i}^{V}\in \mathbb{R}^{d_{\text{model}}\times d_{k}}$.

Finally, the output states are concatenated and projected to produce the final state.
\begin{equation}
\operatorname{MHA}(Q_{drug},K_{omics},V_{omics})=\operatorname{Concat}\left(\text{H}_{1},\ldots\text{H}_{\mathrm{h}}\right)W^{O}
\end{equation}
where $W_{i}^{O}\in \mathbb{R}^{d_{\text{model}}\times d_{k}}$ is the output parameter. In this work we employ $h=2$ parallel attention layers, the dimension of query, key, value are set to 100 and output dimension is $dmodel=d_{k}*h$.

\begin{figure}[!ht]
\centering
\includegraphics[width=1.0\linewidth]{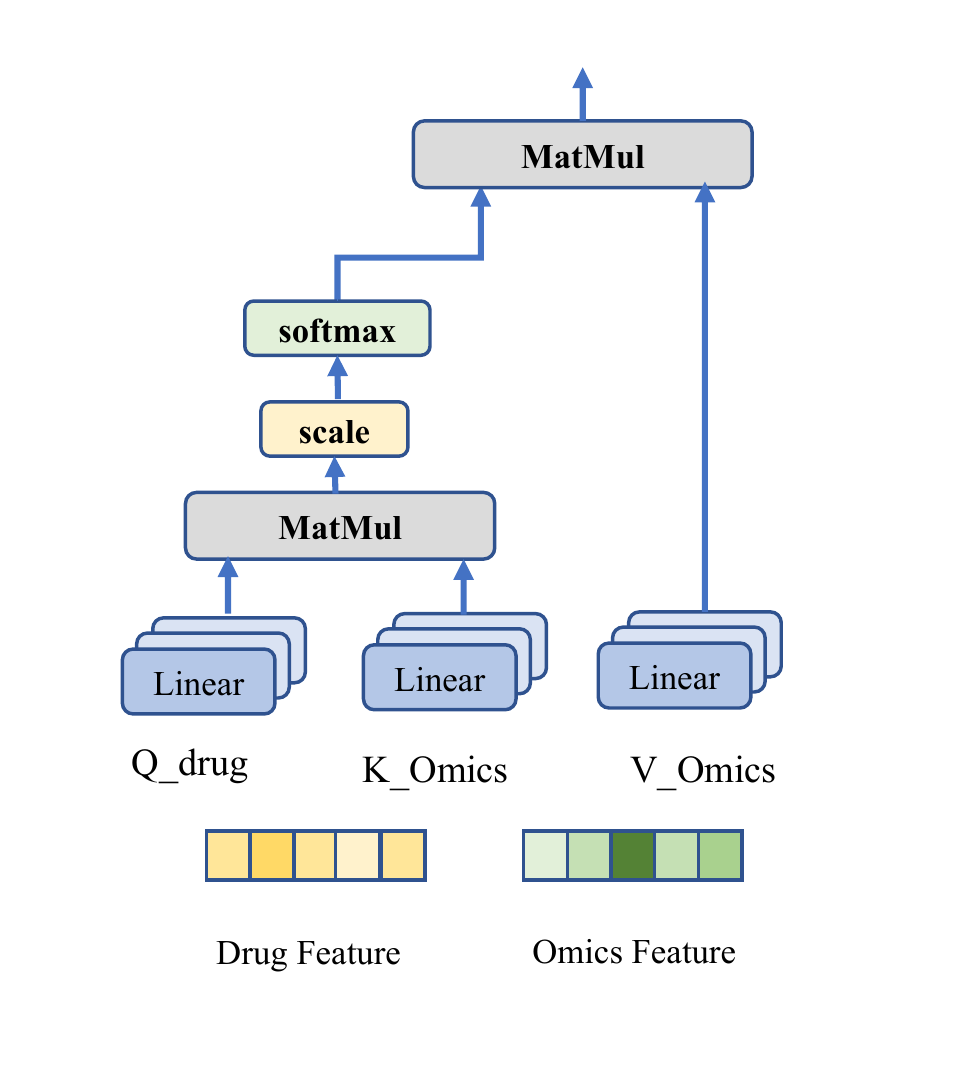}
\caption{The detailed structure of atom omics attention. Every drug atom query pays attention to different genomic, transcriptomic, and epigenomic features, respectively.}
\label{fig:02}
\end{figure}

\subsection{Prediction module}

We applied multi-layer 1D convolutional network to construct the prediction module.
The input of prediction module was the final state features generated by interaction module concatenated with the graph level features of drugs.
The graph level drug features were obtained by global pooling using drug atom features. Batch normalization ~\citep{ioffe2015batch} and dropout~\citep{srivastava2014dropout} were also added after each convolutional layer to alleviate potential overfitting in the training process. We utilized Adam~\citep{kingma2014adam} as the optimizer for updating the parameters of TCR in the back-propagation process. The prediction module finally output natural log-transformed $IC_{50}$ values.

\subsection{Losses}
\label{loss}
In recent years, the ranking loss has been well applied in the recommendation system and shows remarkable performance. Unlike other loss functions, such as Cross-Entropy Loss or Mean Square Error Loss, whose objective is to learn to predict a label directly, the objective of Ranking Losses is to predict relative distances between inputs. 
The cancer drug repurpose problem can be considered as recommending drugs for patients according to the $IC_{50}$ index. Therefore, we assumed that the Dule loss design combing MSE loss and Ranking loss could improve the performance of cancer drug prediction. 
Under the practical situation, we often consider one drug's effectiveness for different people or recommend medications for one person~\citep{morand2021ovarian}. Therefore, we are more interested in the response of samples to the same drug. However, all the existing models treat all samples equally. 

We regarded this problem as a hard example mining problem in machine learning. Many previous works have proved that hard example mining can effectively improve the performance of models in computer vision (CV)~\citep{shrivastava2016training, suh2019stochastic, wang2021attentivenas} and natural language processing (NLP)~\citep{mukherjee2020uncertainty}.
We designed a cross sampling method to mine the 'interested' example from the same drug or same cell line. Thus, the model would pay more attention to the relative distance between same drug or same cell line.

As shown in \textbf{Fig.~\ref{fig:sample}}, we selected pair instances in the same drug (red box) or in the same cell line (green box) to construct training mini-batch. Pairwise ranking loss with margin were used. 

\begin{figure}[!ht]
\centering
\includegraphics[width=0.9\linewidth]{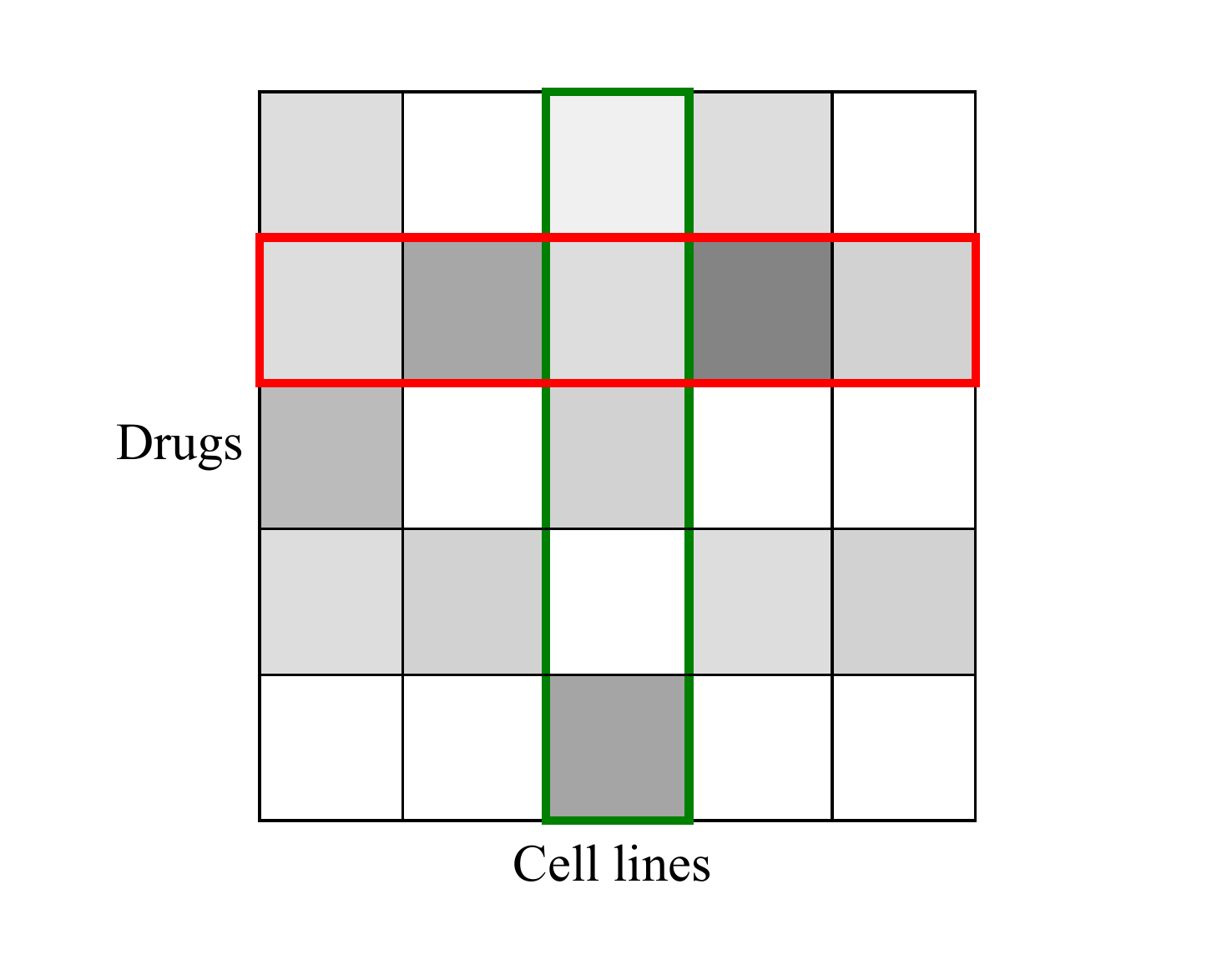}
\caption{Schematic diagram of cross sampling. The pair instances in the same drug (red box) or the same cell line (green box) are used to construct the training batch. }
\label{fig:sample}
\end{figure}

\begin{equation}
L_{rank}(\hat{y}_1,\hat{y}_2,label) = max(0,-label*(\hat{y}_1 - \hat{y}_2) + margin) 
\end{equation}
\begin{equation}
label =\left\{\begin{array}{ccc}
1 & \text { if } &  distance\left(y_{i}, y_{j}\right)>0    \leavevmode \\[\baselineskip]
-1 & \text { if } &  distance\left(y_{i}, y_{j}\right)\le 0
\end{array}\right.
\end{equation}
where y is the ground truth of $IC_{50}$(natural log-transformed). $\hat{y}$ is the prediction of TCR model. In this work the margin is set to 0. The difference of natural log-transformed $IC_{50}$ is used as the distance function of ranking loss.
The overall loss function is shown as below:
\begin{equation}
L=\beta *L_{MSE} + (1-\beta)*L_{rank} \label{equ:loss}
\end{equation}
where $\beta$ is the coefficient hyper-parameter between losses.
The impacts of the coefficient $\beta$ is studied and will be talked in section \ref{sec:drugpcc}. 

\subsection{Baseline methods and evaluation}
The following competing methods were considered. First, traditional machine learning-based methods including RF and Elastic net were considered. For deep learning-based approaches, we chose the baseline models by different technological line: Convolution neural network based tCNNs~\cite{liu2019improving}, Graph neural network-based GraphDRP~\cite{nguyen2021graph}, Hybrid graph convolutional network with multi omics based DeepCDR~\cite{liu2020deepcdr}. The best or default parameters of each method were used for model comparison.

In this work, we predicted natural log-transformed $IC_{50}$ values given the profiles of drugs and cancer cell lines. The common metrics for measuring the statistical correlation between observed values and predicted $IC_{50}$ values are  Pearson’s correlation coefficient (PCC), Spearman’s correlation coefficient (SCC) and root mean squared error (RMSE). 
In clinical scene, we are more interested in the effectiveness of a single drug in different patients. 
Thus, We proposed a new metric called DrugPCC, which is the average of PCC calculated over different drugs. 
Taking the PCC metric as an example, we studied the overall PCC and DrugPCC index's effectiveness in measuring model's performance and will be talked in \ref{sec:drugpcc}. 

To comprehensively evaluate the performance of our model TCR, we demonstrated results under various data splitting settings. We briefly summarized as follows:
\begin{itemize}
    \item \textbf{Predict unknown CDRs.} Random split: This experiment evaluated the performance of models in known drug-cell line pairs. The dataset was shuffled and 80\% were used for training 20\% were used for testing. Five-fold cross-validation was used. 
    \item \textbf{Blind test for both drugs and cell lines.} Leave drug/cell line out: In order to evaluate the predictive power of TCR when given a new drug or new cell line that is not included in the training data. The dataset was randomly splitted into training set (80\%) and test set (20\%) based on the drug or cell line level. Five-fold cross-validation was also used. 
    \item \textbf{External validation.} To evaluate whether TCR can be generalized to other cell line dataset or patient data. We trained TCR model on GDSC dataset and tested on PRISM cell line data and TCGA patient data. 
\end{itemize}

\end{methods}

\section{Results}

\subsection{Model comparison}

We first designed a series of experiments to compare our model with baselines. For this purpose, we collected datasets of the drug and cancer cell line pairs from GDSC~\citep{iorio2016landscape} database and CCLE~\citep{barretina2012cancer} database. We then evaluated the regression performance of TCR and five comparing methods based on the observed $IC_{50}$ values and predicted $IC_{50}$ values.  
We use the average per-drug Pearson's correlation coefficient (DrugPCC), overall Pearson's correlation coefficient (PCC), overall Spearman's correlation coefficient (SCC), and root means square error (RMSE) as metrics.  

First, we followed each paper's setting for the five competitive methods. For ElasticNet and Random Forest, we utilized transcriptomic as input. tCNNs and GraphDRP took SMILES as drug input and genomic mutations as cancer cell input. For DeepCDR and TCR, multi-omics data were used. For fairness, we also compared the performance of different models loading the same omics profiles, which presented in section \ref{sec_ablation}. 

The resluts of different methods of  predicting unknown CDRs under radom split setting are shown in \textbf{Table \ref{Tab:01}}.
Obviously, TCR achieved the best on all four metrics, especially DrugPCC. Generally, deep neural network models significantly outperformed other baselines, since linear or tree-based model may not well capture the structural information within drugs. Among the four deep learning models, TCR outperformed three other deep learning methods with a relatively large margin. TCR achieved a average per-drug Pearson’s correlation (DrugPCC) of 0.685 as compared to 0.642 of DeepCDR, 0.524 of GraphDRP and 0.438 of tCNNs. This conclusion also agrees with other metrics including overall Pearson’s correlation, overall Spearman’s correlation, and root mean square error (RMSE). 

\begin{figure}[h]
    \centering
    \vspace{-0.2cm}
    \subfloat[]{
    \includegraphics[width=3.93cm]{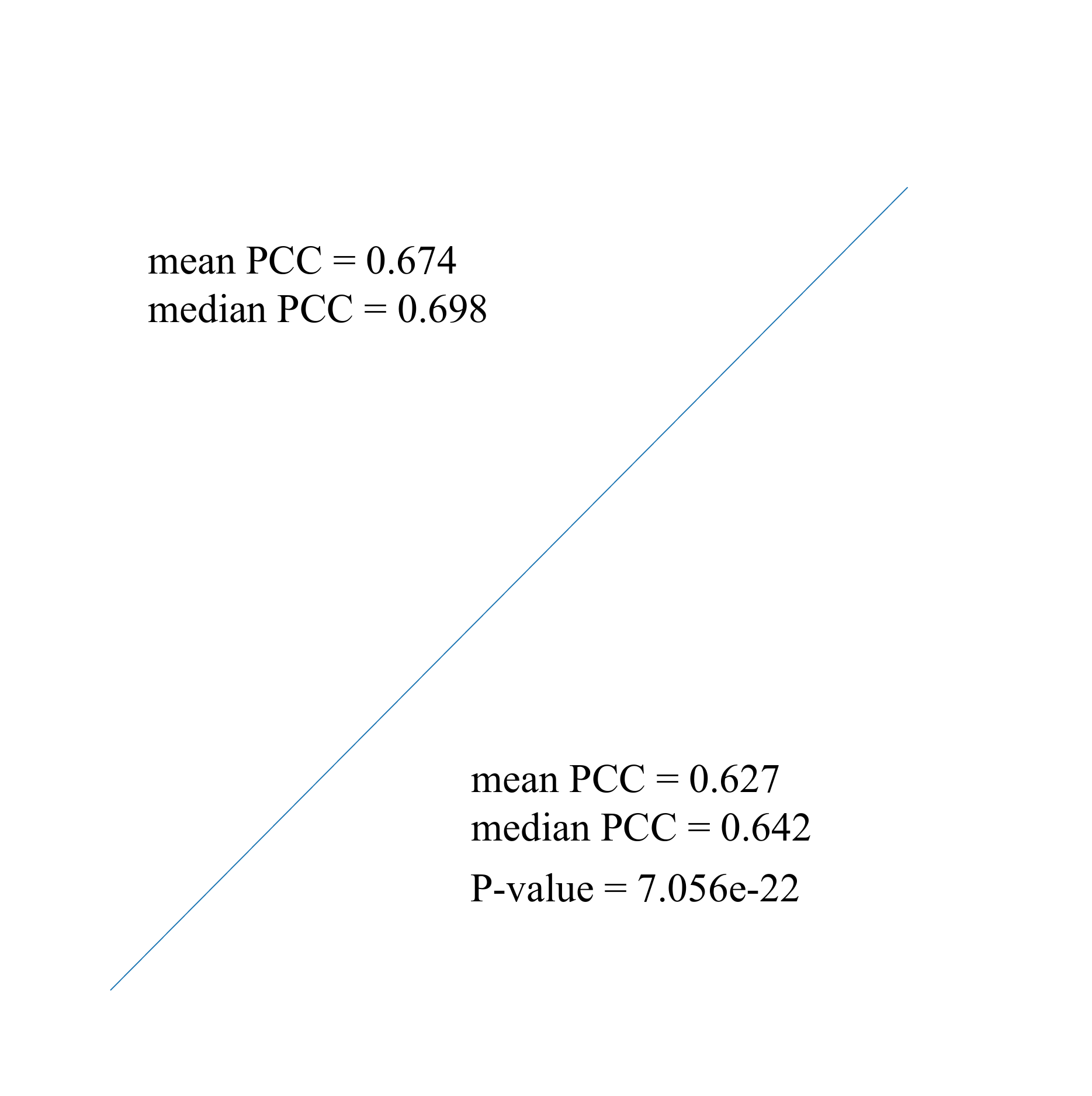}
    
    }
    \subfloat[]{
    \includegraphics[width=3.93cm]{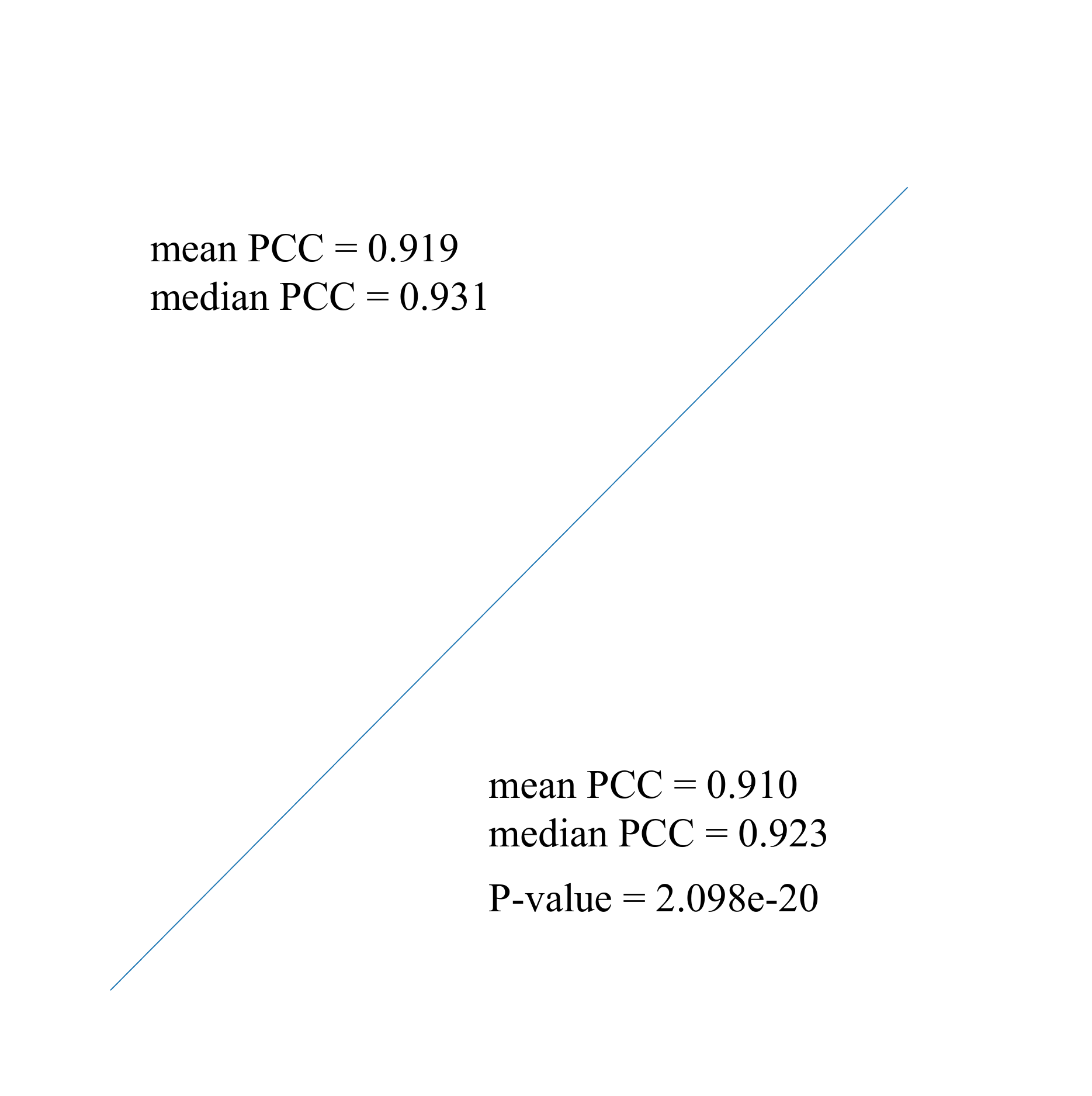}
    }
    
    \quad
    \subfloat[]{
    \includegraphics[width=3.93cm]{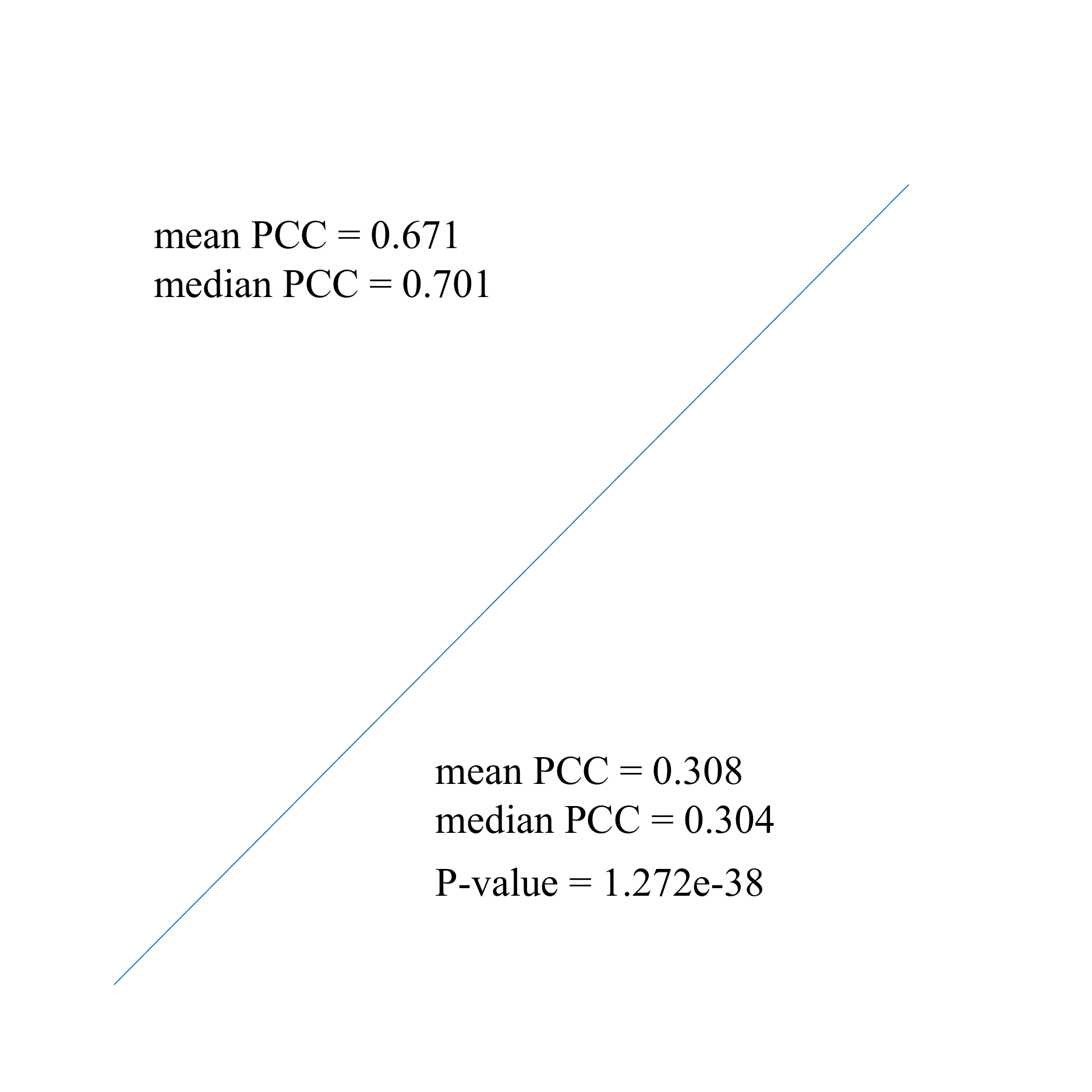}
    }
    \subfloat[]{
    \includegraphics[width=3.93cm]{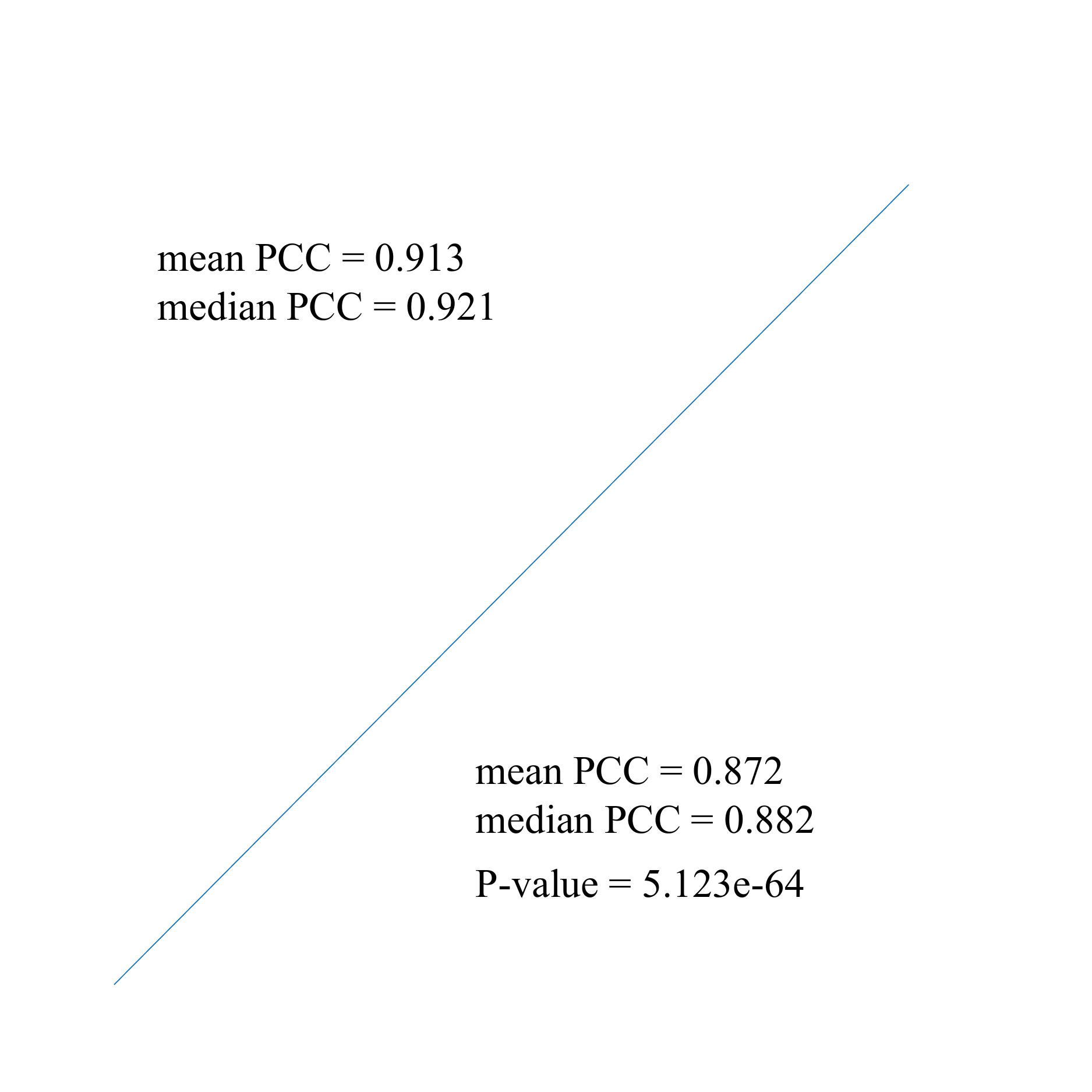}
    }
   \caption{TCR(Ours) and DeepCDR in random splitting tests for drugs are shown. Every dot in the figure (a) and (c) represents a drug, and every dot in the figure (b) and (d) represents a cell line. Figures (a) and (b) highlighted the scatter plots for multi-omics input. Figures (c) and (d) are presented in the case of single RNAseq input. The x-axis and y-axis of each dot represent the Pearson’s correlation of DeepCDR and TCR, respectively. The dot falling into the left upper side in all four figures denotes the case where TCR outperforms DeepCDR. The wilcoxon test P-values are also exhibited in the figure. 
   }
  \label{fig:mix_ccle_drug}
\end{figure}

In particular, we compared our model TCR to the best baseline model DeepCDR in the previous experiments. Wilcoxon rank-sum test were used. In the random splitting setting with multi-omics input, TCR achieved a mean per-drug  Pearson’s correlation (DrugPCC) of 0.674, compared to 0.627 of DeepCDR (\textbf{Fig.~\ref{fig:mix_ccle_drug}} (a), P-value = 7.056e-22). For transcriptomic input only, TCR achieved a mean per-drug  Pearson’s correlation (DrugPCC) of 0.671, compared to 0.308 of DeepCDR (\textbf{Fig.~\ref{fig:mix_ccle_drug}} (c), P-value = 1.272e-38). TCR model achieved an inprovment of 0.363 when transcriptomic data was used independently. 
We also compared the per-cell line Pearson’s correlation with DeepCDR. TCR achieved 0.919 compared to 0.910 of DeepCDR (\textbf{Fig.~\ref{fig:mix_ccle_drug}} (b), P-value = 2.098e-20) for multi-omics input. For transcriptomic input only, TCR achieved a mean per-drug  Pearson’s correlation (DrugPCC) of 0.913, compared to 0.872 of DeepCDR (\textbf{Fig.~\ref{fig:mix_ccle_drug}} (d), P-value = 5.123e-64). 
Compared with multi-omics, TCR model obtains a more significant gain in PCC index than DeepCDR when transcriptomic is used as input~\citep{cui2021ratio}.
\begin{table*}[!ht]
\processtable{Model comparison for random split settings. The overall PCC (PCC), average per-drug PCC for all test drugs (DrugPCC), Spearman's correlation coefficient (SCC), and root means square error (RMSE) for all six methods are presented, and five fold cross-validation is used. Our model TCR outperformed better on all four metrics, especially DrugPCC. 
\label{Tab:01}} {\begin{tabular}{m{3cm} m{2.8cm} m{2.8cm} m{2.8cm} m{2.8cm}}
\toprule Method & PCC & DrugPCC & SCC & MSE\\\midrule
ElasticNet & $0.411 \pm 0.009$ & $0.327 \pm 0.003$ & $0.340 \pm 0.005$ & $6.166 \pm 0.055$\\
RF & $0.750 \pm 0.005$ & $0.330 \pm 0.006$ & $0.670 \pm 0.006$ & $3.242 \pm 0.067$  \\
tCNNs & $0.876 \pm 0.026$ & $0.438 \pm 0.241$ & $0.847 \pm 0.030$ & $2.204 \pm 0.569$  \\
GraphDRP & $0.874 \pm 0.012$ & $0.524 \pm 0.006$ & $0.841 \pm 0.016$ & $2.150 \pm 0.432$  \\
DeepCDR & $0.918 \pm 0.002$ & $0.642 \pm 0.008$ & $0.892 \pm 0.002$ & $1.180 \pm 0.040$  \\
TCR(Ours) & 0.926 $\pm$ 0.002 & $0.685 \pm 0.004$& $0.901 \pm 0.002$ & $1.077 \pm 0.024$\\\botrule
\end{tabular}}{}
\end{table*}


Then, we also compared our model with baselines under the blind test setting. In the previous experiment setting, a drug that appeared in the testing set might also appear in the training set. In this setting, we will predict the response of seen cell lines on unseen drugs (leave drug) or response of novel cell line on seen drugs (leave cell line). 
\textbf{Table \ref{Tab:02}} shows the prediction performance for the blind test with unseen drugs (leave drug out) and unseen cell lines (leave cell line out). We observed that TCR achieved better DrugPCC than other five methods. Particularly, TCR gained a DrugPCC of 0.512 from 5-fold cross-validation, versus 0.503 from the baseline DeepCDR. Compared with traditional machine learning methods (ElasticNet, RF), the four deep learning methods (tCNNs, GraphDRP, DeepCDR, TCR) showed better generalization performance in the leave drug out experimental setting.
Similar to the drug blind test, this experiment evaluated the performance of the model on unseen cell lines. The DrugPCC index is lower than the leave drug settings for all methods, indicating that leave-cell line-out setting is more challenging than leave-drug-out. TCR achieved the best score of 0.365. tCNNs and GraphDRP showed poor generalization ability to new cell lines.

\begin{table*}[!ht]
\processtable{Model comparison for leave drug and leave cell line settings. The average per-drug PCC for all test drugs (DrugPCC) are presented, and five fold cross-validation is used. 
\label{Tab:02}} {\begin{tabular}{m{2.4cm} m{1.9cm} m{1.9cm} m{1.9cm} m{1.9cm} m{1.9cm} m{1.9cm}}
\toprule Experimental setting  & ElasticNet & RF & tCNNs & GraphDRP & DeepCDR & TCR(Ours)\\\midrule
Leave-drug-out & $0.327 \pm 0.025$ & $0.295 \pm 0.045$ & $0.355 \pm 0.200$ & $0.387 \pm 0.187$& $0.503 \pm 0.036$& $0.512 \pm 0.013$\\
Leave-cell line-out & $0.301 \pm 0.073$ & $0.283 \pm 0.065$& $0.018 \pm 0.078$ & $0.029 \pm 0.073$ & $0.312 \pm 0.037$ & $0.365 \pm 0.030$\\\botrule
\end{tabular}}{}
\end{table*}

\subsection{Ablation analysis}\label{sec_ablation}
Since multi-omics data was used as input, we evaluated the contribution of different types of omics data under the random split strategy. For each type of omics data, we discarded other two types of omics data and trained TCR and DeepCDR regression models from scratch for model ablation analysis. As shown in Table \textbf{\ref{Tab:ablation}} , the DrugPCC of TCR ranged from 0.664 to 0.677 using single omic data, demonstrating the contribution of all omics profiles to the model. The genomics data (DNA mutation) contributed the most among different omics profiles for TCR. However, The epigenomics data (DNA methylation) contributed the most among other omics profiles for DeepCDR. For DeepCDR, the DrugPCC varies from 0.298 to 0.645 when using single omic data. TCR model outperformed DeepCDR for all input omics data settings. 


\begin{table}[!ht]
\processtable{Model ablation studies with different experimental settings for TCR(Ours) and DeepCDR. DrugPCC metric is used. 
\label{Tab:ablation}} {\begin{tabular}{m{3.5cm} m{1.9cm} m{1.9cm}}
\toprule Experimental setting  & DeepCDR & TCR(Ours)\\\midrule
Single genomics & $0.644 \pm 0.005$ & $0.677 \pm 0.005$ \\
Single transcriptomics & $0.298 \pm 0.014$ & $0.675 \pm 0.005$\\
Single epigenomics & $0.645 \pm 0.011$ & $0.664 \pm 0.011$\\
Multi-omics   & $0.642 \pm 0.008$ & $0.685 \pm 0.004$\\
\botrule
\end{tabular}}{}
\end{table}

For fairness, we also compared the performance of tCNNs and GraphDRP models which input the genomics (DNA mutation) profiles. tCNNs achieves $0.438 \pm 0.241$ and GraphDRP achieves $0.524 \pm 0.006$ which was inferior to the TCR model ($0.677 \pm 0.005$). Notably, TCR achieved a higher Pearson’s correlation than tCNNs and GraphDRP even with only genomic input. Note that we employed a bottleneck design to adapt the high dimension of genomic features to tCNNs and GraphDRP.  


\subsection{DrugPCC metric and losses}\label{sec:drugpcc}
We studied the overall PCC and DrugPCC index's effectiveness in measuring the model's performance. 
This work used three data splitting methods: random split, leave drug out splitting, and leave cell line out splitting. We replaced the input omics with the one-hot encoding of cell lines and training models in leave cell line out splitting mode. 
As shown in \textbf{Table \ref{Tab:onehot}}, the overall PCC was still higher than 0.8 for all four deep learning methods. This demonstrated that the overall PCC can not well represent the performance of models. However, DrugPCC which calculates and averages the PCC for each drug is consistent with our expectation.

\begin{table*}[!ht]
\processtable{The experiments of DrugPCC and overall PCC over baseline models. We first replaced the input omics feature of cell lines with one-hot encoding. Then the models were trained in the splitting mode of leave cell line.
We observed that the overall PCC is still higher than 0.8 for all four deep learning methods. This illustrated that the overall PCC can not well represent the performance of models. However, DrugPCC, which calculated the PCC for each drug, is consistent with our theoretical expectation.
\label{Tab:onehot}} {\begin{tabular}{m{3cm} m{2.8cm} m{2.8cm} m{2.8cm} m{2.8cm}}
\toprule methods & tCNNs & GraphDRP & DeepCDR & TCR(Ours) \\\midrule
PCC & $0.834\pm0.012$ & $0.831\pm0.013$& $0.830\pm0.009$ & $0.827\pm0.011$ \\
DrugPCC & $0.025\pm0.075$ & $0.029\pm0.038$& $0.027\pm0.024$ & $0.025\pm0.022$\\\botrule
\end{tabular}}{}
\end{table*}
We used ranking loss and MSE loss in this work. To achieve better performance, we conducted a grid search for the hyper-parameter $\beta$ in equation \ref{equ:loss} . From \textbf{Table \ref{Tab:beta}} when $\beta$ is larger than 0.5, our model worked reasonably fine. It indicated that a relatively large MSE loss was beneficial to prediction. TCR worked best when $\beta=0.9$, which was better than $\beta=1$(only MSE loss was used). This showed that ranking loss was also indispensable and had a beneficial effect on CDR.

\begin{table*}[!ht]
\processtable{The table below shows DrugPCC score with different $\beta$ values. It works best when $\beta=0.9$, which is better than $\beta=1$(only MSE loss is used). This shows that rank loss is indispensable and has a beneficial effect on CDR.
\label{Tab:beta}} {\begin{tabular}{m{2.4cm} m{1cm} m{1cm} m{1cm} m{1cm} m{1cm} m{1cm} m{1cm} m{1cm} m{1cm} m{1cm}}
\toprule $\beta$ & 0.1 & 0.2 & 0.3 & 0.4 & 0.5 & 0.6 & 0.7 & 0.8& 0.9 & 1 \\\midrule
DrugPCC & $0.458$ & $0.584$& $0.617$ & $0.636$ &$0.663$ & $0.653$& $0.649$ & $0.651$ & $0.671$ & $0.660$\\\botrule
\end{tabular}}{}
\end{table*}

\subsection{External validation.}

To evaluate whether TCR can be generalized to other independent datasets including cell line and patient data. We trained TCR model on GDSC dataset and tested on PRISM and TCGA dataset.
Because the quantification of whole-genome gene expression levels from primary tumor biopsies is detailed and has been successfully utilized for many years\citep{geeleher2014clinical,geeleher2016cancer}. We used single transcriptomics as the input features of omics. 

\textbf{PRISM Validation.}
First, we trained models on the GDSC dataset and used PRISM as an external validation dataset.
The Pearson’s correlation coefficient of predicted and ground truth of natural log-transformed $IC_{50}$ for each drug was utilized.
In order to identify the performance of different models more clearly, we designed a PCC threshold $p$. After calculating Pearson's correlation coefficient for all drugs, we dropped the drugs if all four methods' Pearson's correlation coefficient is less than $p$. 

\begin{figure}[ht]
\centering
\includegraphics[width=0.95\linewidth]{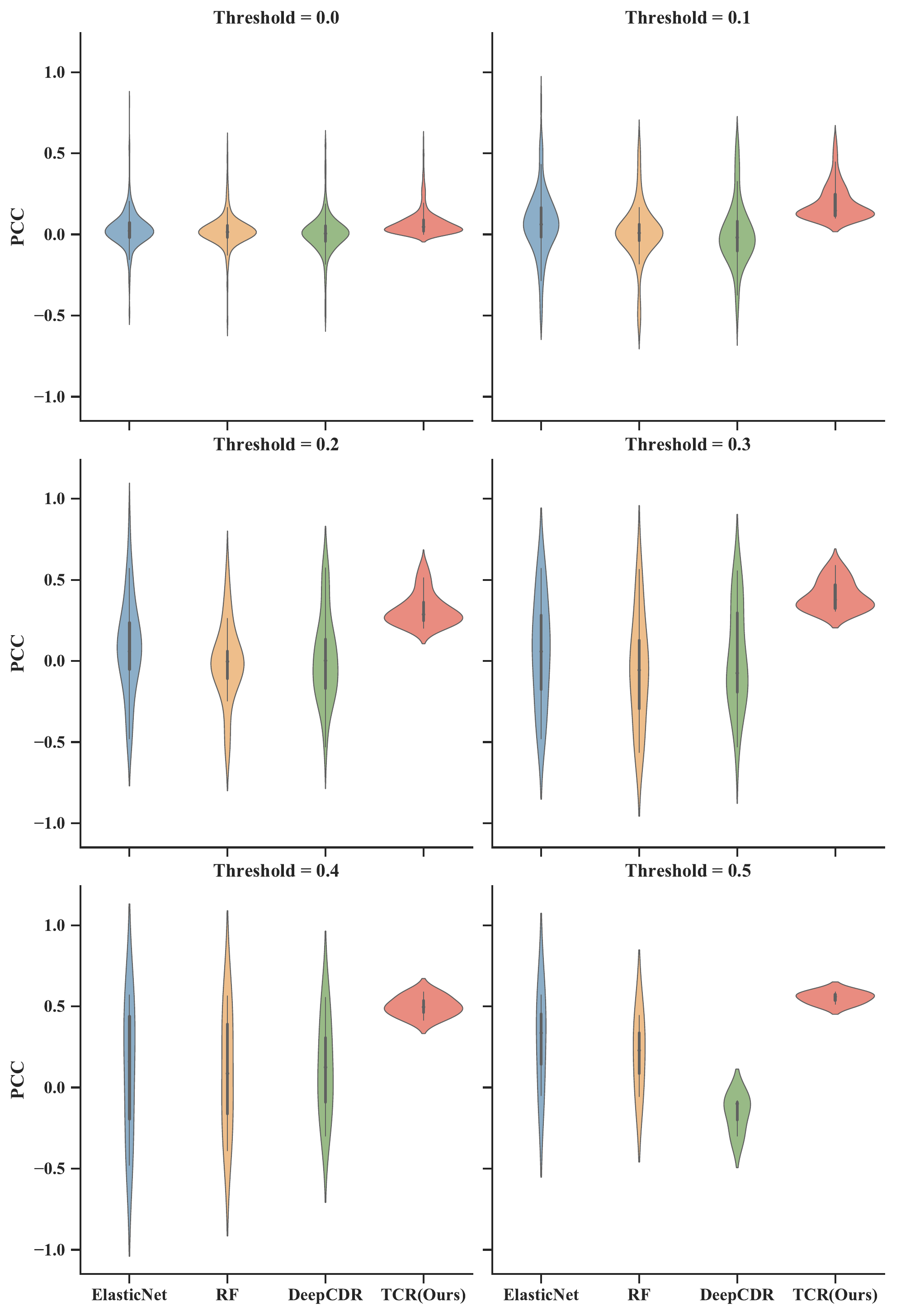}
\caption{PRISM drug screening independate dataset test result of different models. All models used single transcriptomics as the input features of cell line. where p is the threshold of PCC. TCR model obtains higher PCC and minimum variance results compared with other models. }
\label{fig:04}
\end{figure}

We tested results for $p = 0, 0.1, 0.2, 0.3, 0.4, 0.5$. Note that drugs with less than ten samples were also dropped for unreliability due to a small number of samples. The remaining number of drugs were 572, 119, 40, 18, 7, 3 for different thresholds.
\textbf{Fig.\ref{fig:04}}, showed the violin plots of PCC across drugs. Each dot within a violin plot represented the PCC of each drug. 
When p = 0, all four models perform poorly. When we gradually increase the threshold $p$, we could see that the PCC value of the TCR model moves upward and surpasses all other three methods. TCR model also obtained the minimum variance results compared with other models. The results indicated that our model TCR had the potential to predict drug-cell line pairs from noval datasets. 

\begin{figure}[!ht]
\centering
\includegraphics[width=0.9\linewidth]{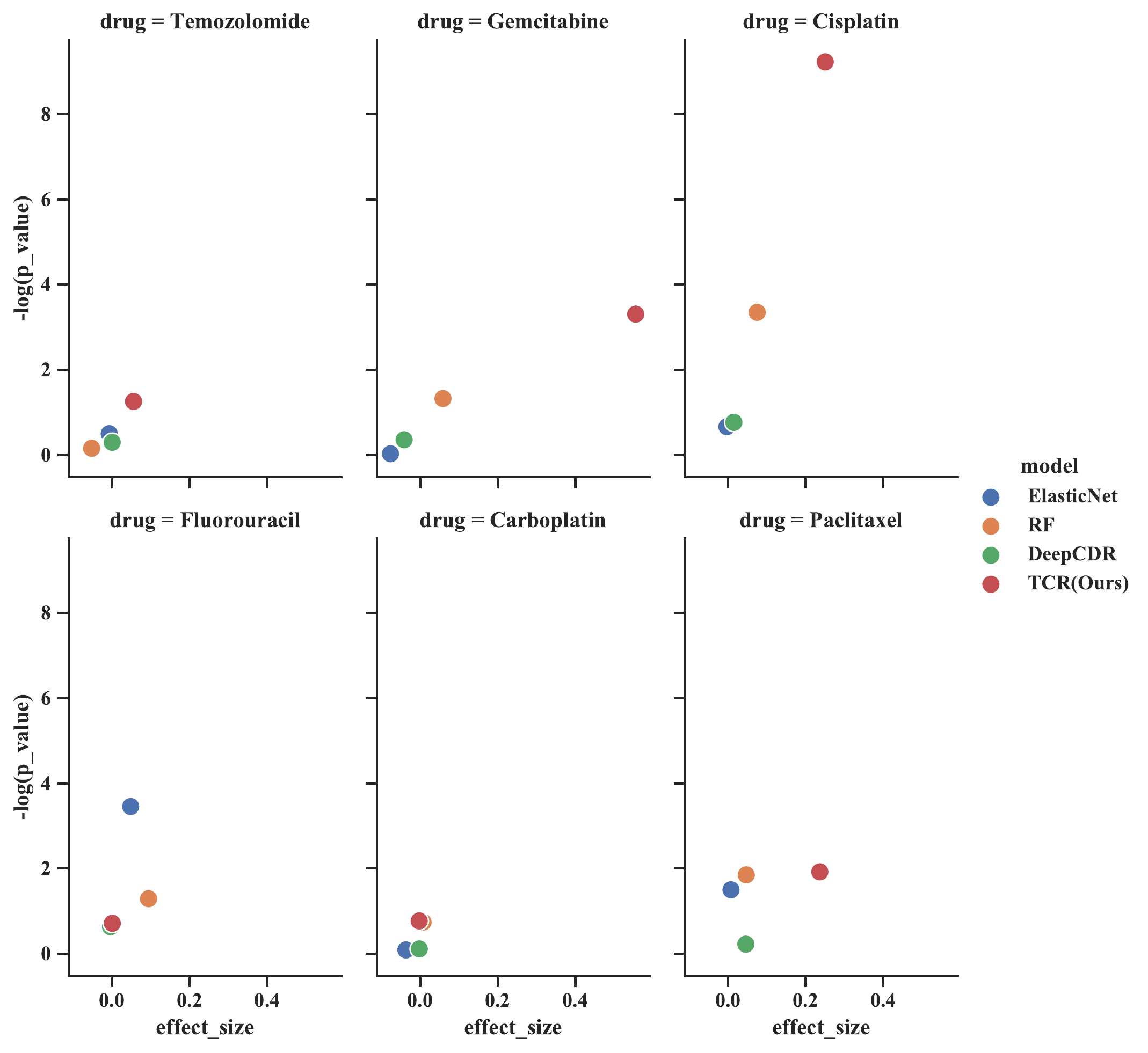}
\caption{The scatter plots of TCGA clinical data for six drugs. We evaluated each model by the effect size measured as the difference in the mean of predicted natural log-transformed $IC_{50}$ values in the responder versus non-responder groups and the associated p values calculated using the paired Wilcoxon test. Overall, TCR model performs better than random forests, elastic nets, and DeepCDR in four drugs (Temozolomide, Gemcitabine, Cisplatin, Carboplatin) by effect size and by statistical significance.}
\label{fig:tcga_pvalue_warp}
\end{figure}

\textbf{TCGA Validation.}
Furthermore, we also evaluated the generalization performance of our model by testing on the patient dataset. 
We utilized a dataset including six drugs (Temozolomide, Gemcitabine, Cisplatin, Fluorouracil, Carboplatin. and Paclitaxel)~\citep{ding2016evaluating}.
The number of samples for each drug is larger than 100. We treated partial and complete response as responder and progressive clinical disease and stable disease as non-responder.

We evaluated each model by the effect size measured as the difference in the mean of predicted natural log-transformed $IC_{50}$ values in the responder versus non-responder groups and the associated P-values calculated using the paired Wilcoxon test. 
The results of our analysis were visualized using scatter plots for each drug in \textbf{Fig.\ref{fig:tcga_pvalue_warp}}. Overall, the TCR model performed better than random forests, elastic nets, and DeepCDR in four drugs (Temozolomide, Gemcitabine, Cisplatin, Paclitaxel) based on effect size and statistical significance. DeepCDR did not yield any significant difference in predicted $IC_{50}$ values between responders and non-responders in all six drugs. For carboplatin, all methods performed poorly, the effect size showed a slight difference in the mean predicted $IC_{50}$ value in the non-responder group compared with the responder group. 
Random forest (RF) offered a promising prediction for cisplatin, since the P-value indicated a significant result. ElasticNet's prediction is more hopeful for Fluorouracil compared with other methods.
\begin{figure}[!ht]
\centering
\includegraphics[width=0.9 \linewidth]{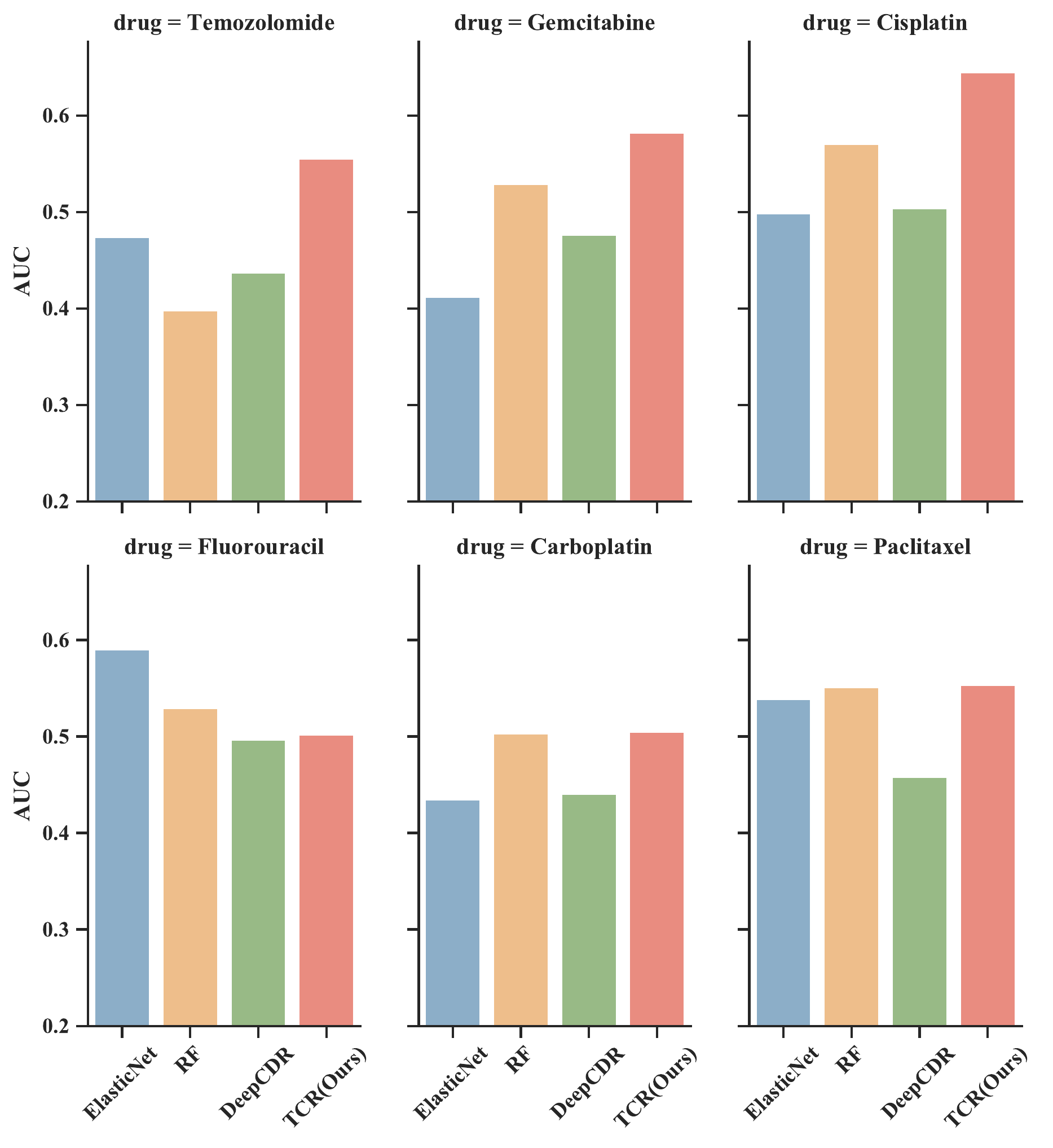}
\caption{TCGA clinical data test for six drugs, the barplots of the AUC values are showed that TCR model achieves higher AUCs compared with other models in three drugs (Temozolomide, Gemcitabine, Cisplatin).}
\label{fig:fig_tcga_auc}
\end{figure}

Finally, we evaluated models based on the metric of AUC (area under the curve). We applied a softmax function to convert the predicted $IC_{50}$ values to binary classification probability.
\textbf{Fig.\ref{fig:fig_tcga_auc}} presented the bar plots of the AUC values for each drug. We found that TCR model yielded statistically significantly higher AUCs compared with other models in three drugs (Temozolomide, Gemcitabine, Cisplatin). This finding was in agreement with the previous conclusions based on the P-value and effect size.

\section{Conclusion and discussion}\label{sec8}

Our study presented a transformer-based TCR model to predict cancer drug response. We found the substructure/atom feature is helpful for cancer drug response prediction compared with the whole drug features. We also found that the transformer network with multi-head atom omics attention is suitable for modeling the interactions of drug atom/substructure and multi-omics data from the perspective of prediction performance. To the best of our knowledge, TCR is the first work that combines GCN and transformer networks to apply for CDR problem. Furthermore, we demonstrated that learning to rank with a cross-sampling strategy helps in the prediction of drug response.
In addition, we found the overall PCC was frequently used in previous works, which may not measure the model's performance correctly. Thus, we proposed a new evaluation metric per-drug Pearson correlation coefficient called DrugPCC, which could better measure the model's performance from the point of practical applications. Based on our findings, we believed that more refined modeling of the interaction among drug substructure and omics data could help to improve CDR prediction accuracy. 
Extensive experiment results from in-vitro cell line data to clinical patient data 
demonstrated TCR is more effective than the current state-of-the-art methods.  This also underlined the predictive capability of TCR and its potential translational value in precise medicine.

However, our methods showed a performance gap between in-vitro data (CCLE) and clinical patient data. It may be caused by the different data distribution of CCLE and TCGA. The lack of data limits the model from being trained directly on the clinical dataset. Additional drug response and multi-omics data will be necessary to train more accurate deep learning models.

We provide two future directions for improving our method. Firstly, we use GCN for the drug feature encoder in this work, which regards drugs as a graph, and the 3D conformation of drugs was ignored. Taking the 3D structure/sub-structure feature of drugs into account may further improve the prediction performance. Secondly, TCR can be leveraged in combination with molecule generation tasks. Current molecule generation models focused on generating general compounds and neglected the profiles of target cancer cells\citep{segler2018planning, sanchez2018inverse,popova2018deep,schneider2020rethinking, wang2022deep}. Compared with previous works, TCR pays attention to the interaction between multi-omics and molecular substructures. The attention score of AOA could be utilized as an index to find the functional part of molecules. Moreover, The prediction of TCR can be used as a reward score for guiding molecule generation for disease-specific novel drug design. 






\section{Competing interests}
There is No Competing Interest.

\section{Author contributions statement}

\section{Acknowledgments}
The authors thank the anonymous reviewers for their valuable suggestions. 

\bibliographystyle{abbrvnat}
\bibliography{Ref}




\end{document}